# OD-GCN: OBJECT DETECTION BOOSTED BY KNOWLEDGE GCN

*Zheng Liu*[1]*, Zidong Jiang*[1]*, Wei Feng*[1]*, Hui Feng*[2]

[1]iQIYI Inc, China
{liuzheng05, jiangzidong, frankfengw}@qiyi.com
[2]Fudan University, Shanghai 200433, China
hfeng@fudan.edu.cn

## ABSTRACT

Classical CNN based object detection methods only extract the objects' image features, but do not consider the high-level relationship among objects in context. In this article, the graph convolutional networks (GCN) is integrated into the object detection framework to exploit the benefit of category relationship among objects, which is able to provide extra confidence for any pre-trained object detection model in our framework. In experiments, we test several popular base detection models on COCO dataset. The results show promising improvement on mAP by 1~5pp. In addition, visualized analysis reveals the benchmark improvement is quite reasonable in human's opinion.

## 1. INTRODUCTION

Humans have a talent for object detection for many reasons. For one reason, humans can recognize an object not only by knowledge of the object itself, but also by its surrounding objects. For the example shown in Fig.1, there is a detection result of desert with 0.99 confidence, while there is another possible result with 0.34 for fish and 0.25 for lizard. It is highly unlikely for a fish to appear on desert. So it is reasonable to reduce the confidence level for a fish detection, and raise that for lizard. This example shows that the detection task can be improved from the high-level knowledge in the environment.

Compared with human, most object detection networks lack of category relationship knowledge. Deep learning object detection networks like Faster R-CNN, SSD and YOLO [1-3], have similar ways for detection. These models locate an object on an image, then crop it (implicitly or explicitly) and classify it by the cropped part. This detection procedure performs well on many benchmark datasets, but it may achieve better performances with other knowledge. So there comes a widely discussed issue about knowledge information. Graph convolution networks (GCN) [23-24] is a powerful tool to describe such irregular-structured information.

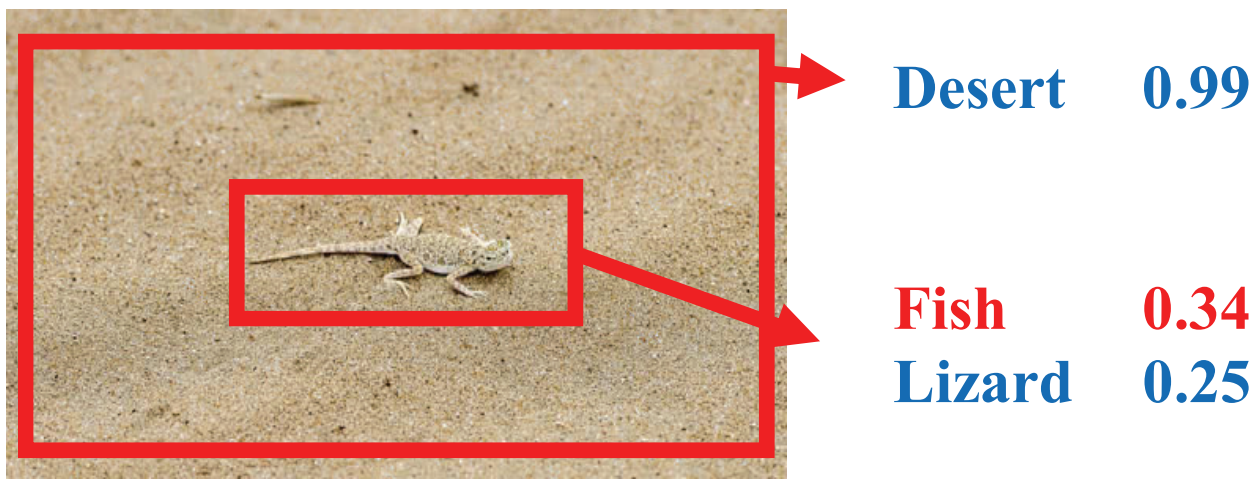

**Fig.1.** A deficiency for classical object detection framework. When desert is detected with a high confidence level like 0.99, this information should help to classify other confusing objects. Besides, the confidences of fish and lizard are very close, it should be able to adjust the original false result to a true one.

In this paper, we propose a new object detection framework called OD-GCN (object detection with graph convolutional network) to boost the detection performance. OD-GCN contains two parts, OD model and GCN model. OD model can be any classical object detection model like SSD or Faster RCNN. GCN model can utilize knowledge graph for post-processing, such as parsing the Wikipedia to build a category relationship knowledge graph. We follow the methods [21] to build the graph with the conditional probability between every two categories in MSCOCO [4] dataset. The GCN's input is OD model's output. After graph convolution operation, the final output should be much more reasonable and accurate.

There are three main contributions in this paper. Firstly, we propose a new framework for object detection, which can improve classical models by knowledge graph with GCN. Although GCN is already used in some other computer vision tasks, this paper is the first work to bring knowledge GCN into detection field as far as we know. Secondly, we improve the GCN with an adaptive parameter representing the weight of neighbor categories' information. Finally, our framework is verified to improve the performance of many classical detection models in experiments.

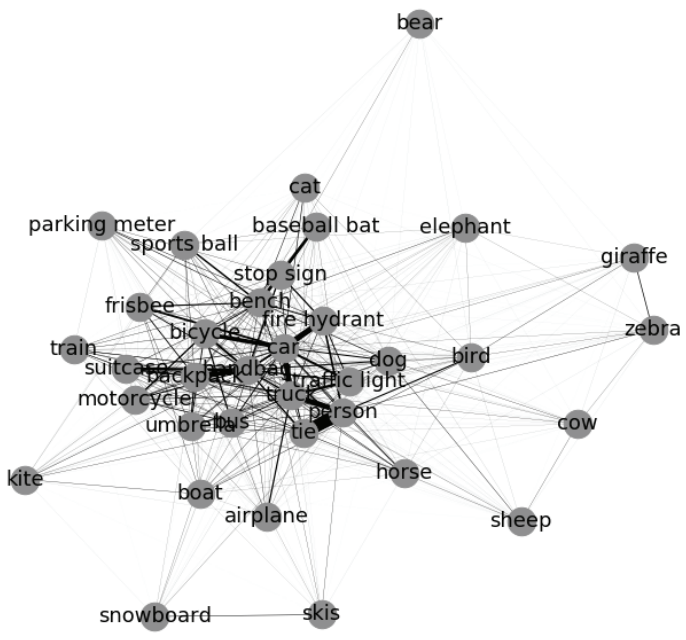

**Fig.2.** Part of the COCO knowledge graph. The knowledge graph represents the closeness of the relationship between each two categories. The thicker the edge line is, the closer relationship between the two connected categories.

## 2. RELATED WORK

Object detection is an essential task in computer vision field. It is backbone for many advanced applications, such as facial detection, autonomous driving, drone photography. With help of well-annotated datasets like ImageNet [9], PASCAL VOC [9], MSCOCO [4] and the famous CNN methods like SSD, YOLO, Faster-RCNN [1-3], object detection is able to perform quite well. Nowadays, researchers keep seeking ways to enhance detection accuracy.

**Image Classification with Extra Information.** Object detection can be divided into two main parts: location regression and image classification. Many new classification frameworks are trying to utilize extra information beyond image itself.

Firstly, category's semantic embedding is one kind of well-known extra information. Socher *et al.* [11] trained a network for image and language, learning mapping between image representations and word embedding. Frome *et al.* [18] inspired by Socher, proposed DeViSE system to map image and text with a good performance on object classification tasks. Norouzi *et al.* [19] proposed a similar system called ConSE, combining ConvNet features and word embedding together. Changpinyo *et al.* [20] proposed a zero-shot classifier aligning sematic and visual information.

Secondly, knowledge graph is also widely used in classification field. Li *et al.* [15] improved performance of multi-label classification with probability relationship knowledge graph. Lee *et al.* [17] proposed ML-ZSL framework, developing the relationship knowledge graph for zero-shot classification.

**GCN and Image Classification.** Nowadays, GCN becomes quite a popular tool to deal with knowledge graph. Naturally, GCN is used in the image classification field combined with knowledge graph. Chen *et al.* [21] proposed a GCN-based framework called ML-GCN for multi-label classification, with sematic embedding as GCN's input. Wang *et al.* [14] transfer the previous idea for zero-shot classification tasks, proposing a GCN network through a WordNet knowledge graph. Zhang *et al.* [22] combined GCN, sematic embedding, and probability knowledge together, building a sematic knowledge graph and a scene probability graph to refine traditional classification results.

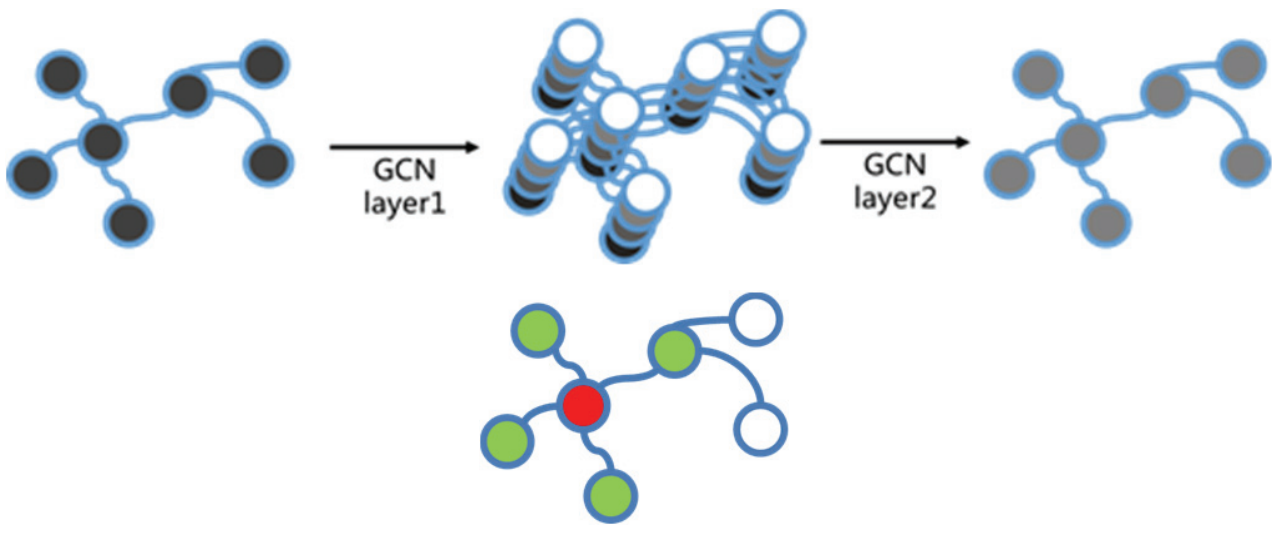

**Fig.3.** The illustration of our GCN model. After the two graph convolutional layers, the channels are 1→4→1 from $\bar{\boldsymbol{G}}$ to $\hat{\boldsymbol{G}}$. New features of each node are aggregated from its neighbor nodes.

Inspired by above works in classification field, we bring GCN and knowledge graph into object detection field. Differently, OD model's output is chosen as the GCN's input, instead of sematic embedding. So the GCN model has stronger connection with image content. Besides, we add a trainable adaptive parameter in GCN to prove the effectiveness of importing the knowledge graph.

## 3. APPROACH

Our key idea is utilizing information of objects' relationship for object detection task. In following parts, we will firstly introduce the details of knowledge graph and GCN structure. Then we will explain the whole OD-GCN framework and how it works.

### 3.1. Knowledge Graph for OD-GCN

The category relationship knowledge graph is built following the way in ML-GCN [21] with COCO 2014 training dataset. For the established COCO knowledge graph, each node represents each category and the graph edge from Node $A$ to Node $B$ is calculated by the conditional probability $P(B|A)$. For instance, if cat and dog appear together 4 times in COCO dataset, and cat appears 8 times totally in the dataset, the edge from cat to dog is defined as $\mathrm{P}(dog|cat)$=4/8=0.5. Part of the COCO knowledge graph has been visualized in Fig.2. In addition, some other category knowledge graphs built up by category correlation matrix are also capable for our OD-GCN framework.

After the preparation of the knowledge graph, the adjacent matrix $\boldsymbol{A} \in \mathbb{R}^{C\times C}$, and the degree matrix $\boldsymbol{D} \in \mathbb{R}^{C\times C}$ can be computed. $C$ is the number of categories, which is 91 in our graph, including the background. The conditional probabilities between background and other categories are set to be zero. In the following part, we will introduce the overall framework for object detection with GCN based on COCO knowledge graph.

### 3.2. Graph Convolutional Network for OD-GCN

Graph Convolutional Network is a kind of deep trainable network designed for graph structure. Graph convolution on a graph is quite similar to classic 2D convolution on an image. The new features on a node is related to previous features of its neighbor nodes. The sketch of our graph convolution layers is shown in Fig.3.

There are various graph convolution functions for a graph convolution layer. In this paper, the complete function of a graph convolution layer is adaptive, defined as

$$\boldsymbol{H}_{l+1} = ReLU((\alpha\boldsymbol{D}^{-1}\mathbf{A} + \boldsymbol{I})\boldsymbol{H}_l\boldsymbol{W}_l) + \boldsymbol{B}_l, \tag{1}$$

where $\boldsymbol{D}^{-1}$ can normalize the adjacent matrix by the number of neighbors. Otherwise, nodes with too many neighbors will overwhelm other useful features. $\boldsymbol{I}$ is the identity matrix to remain features of every node itself. The input and output features are defined as $\boldsymbol{H}_l \in \mathbb{R}^{C\times C1}$, $\boldsymbol{H}_{l+1} \in \mathbb{R}^{C\times C2}$, where $C1$ and $C2$ are the input and output channel number. The trainable weight matrix is $\boldsymbol{W}_l \in \mathbb{R}^{C1\times C2}$, which is initialized to a matrix full of ones. This initial value is helpful for quick convergence when training. $\boldsymbol{B}_l \in \mathbb{R}^{C\times C2}$ is a trainable variable to control the output's mean value. The initial value of $\boldsymbol{B}_l$ is set to be 1.0. After one graph convolution layer, the channel number $C1$ can be turned to $C2$. The subscript $l$ means these parameters belong to the $l^{th}$ layer.

We introduce the adaptive parameter $\alpha$ into the graph convolution operation. This trainable parameter is used to control the ratio of a category itself and its neighbors. If $\alpha$ is 0, it means that GCN do not need the neighbors' information. In experiments, $\alpha$ is usually 0.1~0.5, which proves that neighbors' information is used and the knowledge graph really works for our task.

### 3.3. Overall Object Detection with GCN

The overall framework is displayed in Fig.4. OD-GCN includes two main parts, a well-trained object detection stage and a GCN post-processing stage based on category relationship knowledge graph. Note that it is total free to use any CNN detection model as long as the model is well-trained.

*3.3.1. Classical object detection stage*

Any well-trained classical detection framework is capable on this stage. In our experiment, we choose five SSD and three Faster R-CNN base models. Processed by the base detection model, the input image will be converted to a raw confidence matrix $\overline{\boldsymbol{Y}} \in \mathbb{R}^{B\times C}$ , where $B$ is the number of all the detected boxes (after NMS or other similar post-processing operations). $\overline{\boldsymbol{Y}}_{bc}$ represents the probability of $c^{th}$ category of the $b^{th}$ box. The raw confidence matrix should be processed by a Softmax layer. For each detection box, $\sum_{c=1}^{C}\overline{\boldsymbol{Y}}_{bc} = 1$.

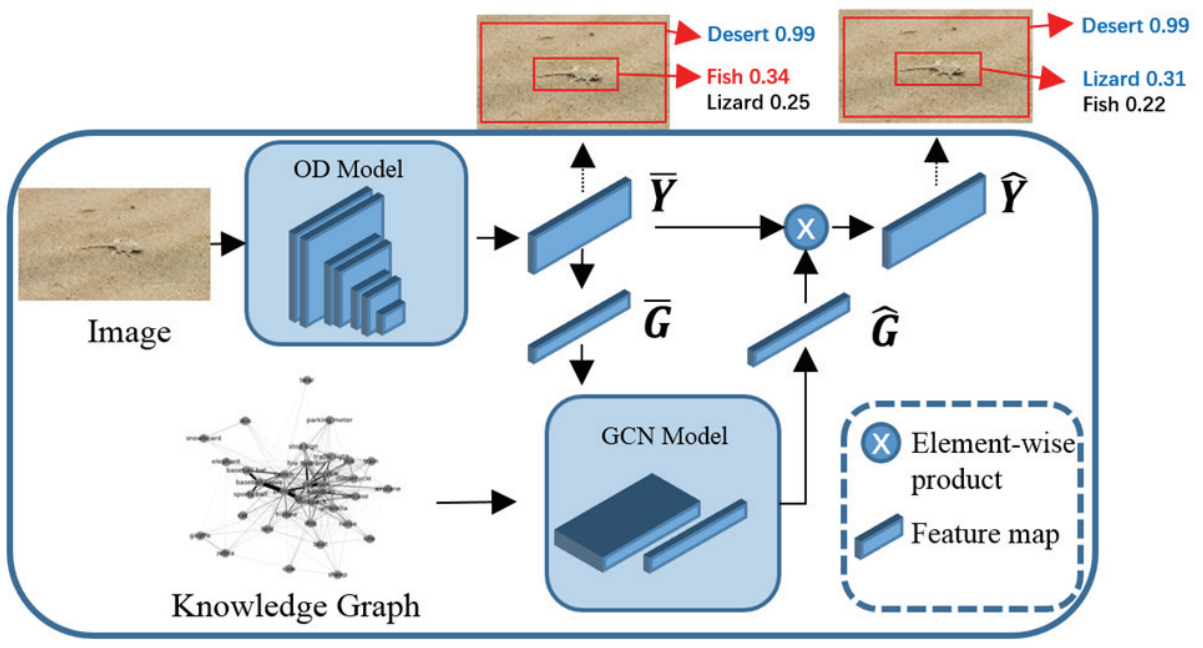

**Fig.4.** A brief procedure of OD-GCN framework.

*3.3.2. GCN post-processing stage*

A category relationship knowledge graph has been set up by COCO dataset in the previous part. Besides, the max value on box axis of the $B\times C$ raw confidence matrix $\overline{\boldsymbol{Y}}$ is remained. This output vector $\overline{\boldsymbol{G}} \in \mathbb{R}^{C}$ is the input feature of our GCN post-processing framework. The formula is,

$$\overline{\boldsymbol{G}}_c = max_{b=1,2,\ldots B}(\overline{\boldsymbol{Y}}_{bc}), c = 1,2,\ldots,C, \tag{2}$$

where $\overline{\boldsymbol{G}}_c$ is the value of $c^{th}$ category for GCN's input vector $\overline{\boldsymbol{G}}$, the max confidence score of $c^{th}$ category among all detection boxes.

Why is max value chosen as GCN's input feature? Mean value or sum value has also been thought about. However, both sum and mean value are easy to be affected by number of total boxes. Max value is much more stable, and correctly reflect the confidence of "there is category $A$ in the image". And the experiments result also show that max value is better for our framework.

The whole GCN model is defined as $GCN()$, including several graph convolutional layers. After the GCN processing, $\widehat{\boldsymbol{G}} = GCN(\overline{\boldsymbol{G}})$. $\widehat{\boldsymbol{G}} \in \mathbb{R}^{C}$ represents the weight for confidence adjustment. And the final confidence matrix $\widehat{\boldsymbol{Y}} \in \mathbb{R}^{B\times C}$ is the element-wise product between raw confidence matrix $\overline{\boldsymbol{Y}}$ and adjustment weight $\widehat{\boldsymbol{G}}$,

$$\widehat{\boldsymbol{Y}} = \overline{\boldsymbol{Y}} \odot \widehat{\boldsymbol{G}} = \begin{bmatrix} \bar{y}_{11} & \cdots & \bar{y}_{1C} \\ \cdots & \cdots & \cdots \\ \bar{y}_{B1} & \cdots & \bar{y}_{BC} \end{bmatrix} \odot [\hat{g}_1 \quad \cdots \quad \hat{g}_C]$$

$$= \begin{bmatrix} \bar{y}_{11} * \hat{g}_1 & \cdots & \bar{y}_{1C} * \hat{g}_C \\ \cdots & \cdots & \cdots \\ \bar{y}_{B1} * \hat{g}_1 & \cdots & \bar{y}_{BC} * \hat{g}_C \end{bmatrix}, \tag{3}$$

where $\hat{y}_{bc}$ represents the final probability of $c^{th}$ category of the $b^{th}$ box. Note that if the vector $\widehat{\boldsymbol{G}}$ is a vector full of ones, it means the raw confidence matrix do not need any adjustment by knowledge graph.

The overall loss function is defined as cross entropy,

$$L = -\frac{1}{B}\sum_{b=1}^{B}\sum_{c=1}^{C} y_{bc}\log(\hat{y}_{bc}), \tag{4}$$

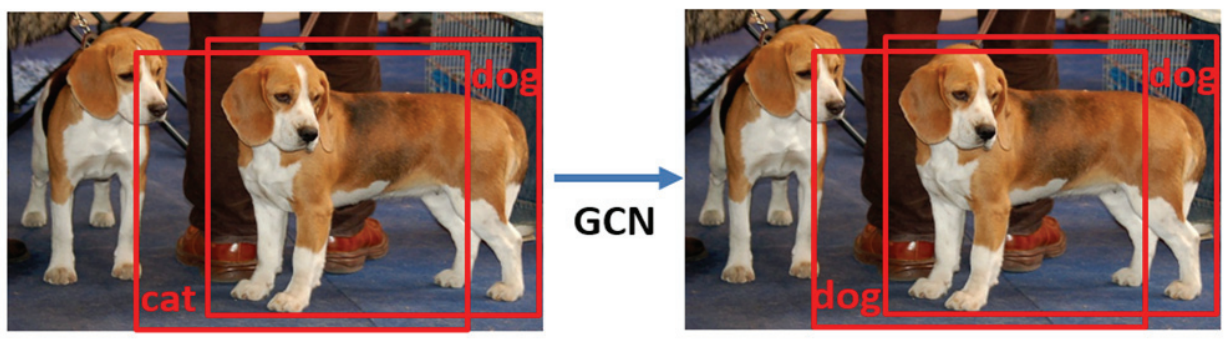

**Fig.5.** One example for explanation of the need of NMS after OD-GCN. Although the base model has NMS operation, it is possible for GCN to output some boxes of the same category with really high IoU score.

where $y_{bc} \in \{0,1\}$ is the ground-truth label for the $c^{th}$ category in $b^{th}$ box. The labels can be calculated by the maximal IoU with the ground-truth boxes in the dataset. With this loss function, the weight matrix can be learnt by training.

### 3.4. Implementation details

We only need to train the GCN part in our framework. According to previous studies, GCN is easy to overfit, so many researchers choose GCN within 6 layers for adjustment [14, 21]. Our GCN is composed of two graph convolutional layers with channel numbers as 1 → 4 → 1. The shapes of feature maps are (1, 91) → (4, 91) → (1, 91) from $\bar{\boldsymbol{G}}$ to $\hat{\boldsymbol{G}}$. The activation function is *ReLU*, while the optimizer is Adam. The initial learning rate is 0.01, and the learning rate will decay to 95% after every 100 training steps.

To accelerate the training process, the regularization term is added to the final loss function. The loss function is,

$$L = -\sum_{c=1}^{C} \hat{\boldsymbol{y}}_c \log(\boldsymbol{y}_c) + \beta \sum_{\hat{g}_c \epsilon \hat{\boldsymbol{G}}} (\hat{g}_c - 1)^2, \quad (5)$$

where $\beta \sum_{\hat{g}_c \epsilon \hat{\boldsymbol{G}}} (\hat{g}_c - 1)^2$ is the regularization term of $\hat{\boldsymbol{G}}$ which restricts $\hat{\boldsymbol{G}}$ nearly to full of ones. The regularization term can accelerate training by stabilizing the GCN output range. In our experiment, $\beta$ is set to be 1.0.

In addition, it is necessary for OD-GCN to have a non-maximum suppression (NMS) post-processing operation. Although most base detection frameworks like SSD and Faster R-CNN already have a NMS post-processing procedure, it is possible for OD-GCN to output two boxes of the same class with high IoU score. One example is shown in Fig.5.

## 4. EXPERIMENT

In this part, we will present our experiments on several different SSD and Faster R-CNN models trained by Google on MSCOCO 2014 Minival dataset.

Microsoft COCO is a well-known benchmark dataset for instance segmentation and object detection. COCO 2014 dataset contains 82,081 images as the training set, and 40,504 images as the validation set. The minival dataset in this paper is selected by Google from the original 40,504 validation set, which contains 8,060 validation images. The detailed image list and pre-trained models can be found on the Github homepage of Tensorflow, Google [25].

**Table.1.** mAP of different models with OD-GCN optimization. We test eight models. It includes
1) ssd_mobilenet_v1_0.75_depth, 2) ssd_mobilenet_v1,
3) ssd_mobilenet_v1_fpn, 4) ssd_mobilenet_v2,
5) ssd_resnet50_v1_fpn, 6) faster_rcnn_resnet50,
7) faster_rcnn_resnet101, 8) faster_rcnn_inception_v2.
The last four summarized results SSD, ODGCN(SSD), Faster RCNN, ODGCN(Faster RCNN) are the average values of corresponding values above.

| Model | | | mAP | mAP @0.5 | mAP @0.75 |
|---|---|---|---|---|---|
| SSD | 1) | | 6.15 | 12.4 | 4.53 |
| | | +ODGCN | 9.28 | 17.5 | 7.31 |
| | 2) | | 29.9 | 48.5 | 27.2 |
| | | +ODGCN | 33.9 | 52.3 | 32.1 |
| | 3) | | 23.2 | 36.9 | 22.0 |
| | | +ODGCN | 26.3 | 40.6 | 25.5 |
| | 4) | | 29.3 | 48.0 | 26.3 |
| | | +ODGCN | 32.0 | 50.7 | 29.7 |
| | 5) | | 24.0 | 37.5 | 22.9 |
| | | +ODGCN | 27.6 | 41.9 | 26.7 |
| Faster RCNN | 6) | | 22.6 | 38.5 | 20.1 |
| | | +ODGCN | 23.6 | 39.7 | 21.4 |
| | 7) | | 24.0 | 38.8 | 22.2 |
| | | +ODGCN | 25.2 | 40.3 | 23.6 |
| | 8) | | 21.8 | 37.7 | 19.2 |
| | | +ODGCN | 23.2 | 39.4 | 20.7 |
| SSD | | | 22.5 | 36.6 | 20.5 |
| ODGCN(SSD) | | | **25.8** | **40.6** | **24.2** |
| Faster RCNN | | | 22.8 | 38.3 | 20.5 |
| ODGCN(Faster RCNN) | | | **24.0** | **39.8** | **21.9** |

In this paper, our OD-GCN framework has been tested on five SSD models and three Faster R-CNN models. We report the evaluation results of average precision (AP) and mean average precision (mAP). We choose the three benchmark mAP evaluation metrics for object detection tasks, mAP@0.5, mAP@0.7, mAP. In addition, we remove boxes with max confidence lower than 0.1, instead of 0.5 to focus on more confusing raw detection results. The mAP result of our experiment is shown in Table.1.

In experiment, OD-GCN improves the mAP metrics on different base object detection models by about 1~5pp. This shows the effectiveness of our OD-GCN framework for post-processing the classical object detection networks. Detailed AP evaluation scores of all categories are shown in Fig.6 and

**Table.2.** Average precision (AP)@0.5 on some categories of COCO dataset. The last two summarized results SSD, ODGCN(SSD) are the average values of corresponding values above.

| Model | car | bus | train | truck | boat | bench | bird | cat | dog | horse | sheep | zebra | clock | vase | sink |
|---|---|---|---|---|---|---|---|---|---|---|---|---|---|---|---|
| 1) | 8.3 | 15.9 | 25.3 | 9.4 | 8.8 | 9.6 | 10.4 | 24.3 | 18.2 | 18.1 | 15.0 | 29.1 | 15.4 | 9.2 | 12.2 |
| 1)+ODGCN | 12.3 | 25.9 | 26.8 | 19.9 | 10.5 | 14.2 | 17.1 | 36.9 | 42.7 | 30.5 | 19.9 | 29.0 | 19.8 | 13.3 | 12.7 |
| 2) | 31.4 | 56.1 | 74.2 | 57.6 | 35.8 | 51.2 | 43.4 | 71.7 | 46.7 | 54.3 | 47.4 | 67.9 | 55.0 | 50.9 | 55.8 |
| 2)+ODGCN | 33.3 | 62.9 | 77.7 | 67.5 | 39.0 | 56.1 | 56.6 | 80.5 | 56.8 | 60.8 | 50.1 | 69.8 | 63.9 | 60.5 | 67.3 |
| 3) | 41.6 | 41.9 | 49.3 | 30.8 | 36.8 | 32.2 | 39.3 | 45.4 | 43.3 | 40.8 | 42.8 | 54.9 | 52.2 | 36.6 | 41.8 |
| 3)+ODGCN | 45.3 | 51.9 | 55.7 | 41.3 | 42.8 | 38.7 | 45.1 | 52.6 | 59.6 | 50.2 | 50.0 | 54.9 | 54.1 | 40.4 | 48.2 |
| 4) | 35.9 | 61.2 | 73.6 | 49.9 | 35.4 | 51.8 | 37.2 | 67.8 | 64.8 | 55.9 | 42.5 | 69.5 | 57.7 | 45.7 | 51.7 |
| 4)+ODGCN | 37.3 | 66.6 | 78.0 | 63.3 | 36.6 | 59.4 | 50.8 | 74.8 | 72.6 | 64.9 | 47.6 | 71.6 | 65.5 | 54.9 | 63.2 |
| 5) | 42.9 | 39.4 | 50.9 | 31.6 | 35.9 | 31.0 | 39.3 | 48.3 | 47.5 | 47.2 | 43.4 | 53.3 | 51.9 | 36.2 | 39.5 |
| 5)+ODGCN | 46.5 | 49.3 | 55.7 | 42.9 | 39.6 | 35.3 | 44.2 | 53.4 | 59.7 | 55.7 | 48.5 | 52.6 | 53.4 | 40.9 | 42.5 |
| SSD | 32.0 | 42.9 | 54.6 | 35.8 | 30.5 | 35.1 | 33.9 | 51.5 | 44.1 | 43.2 | 38.2 | 54.9 | 46.4 | 35.7 | 40.2 |
| ODGCN(SSD) | **34.9** | **51.3** | **58.7** | **46.9** | **33.7** | **40.7** | **42.7** | **59.6** | **58.2** | **52.4** | **43.2** | **55.5** | **51.3** | **42.0** | **46.7** |

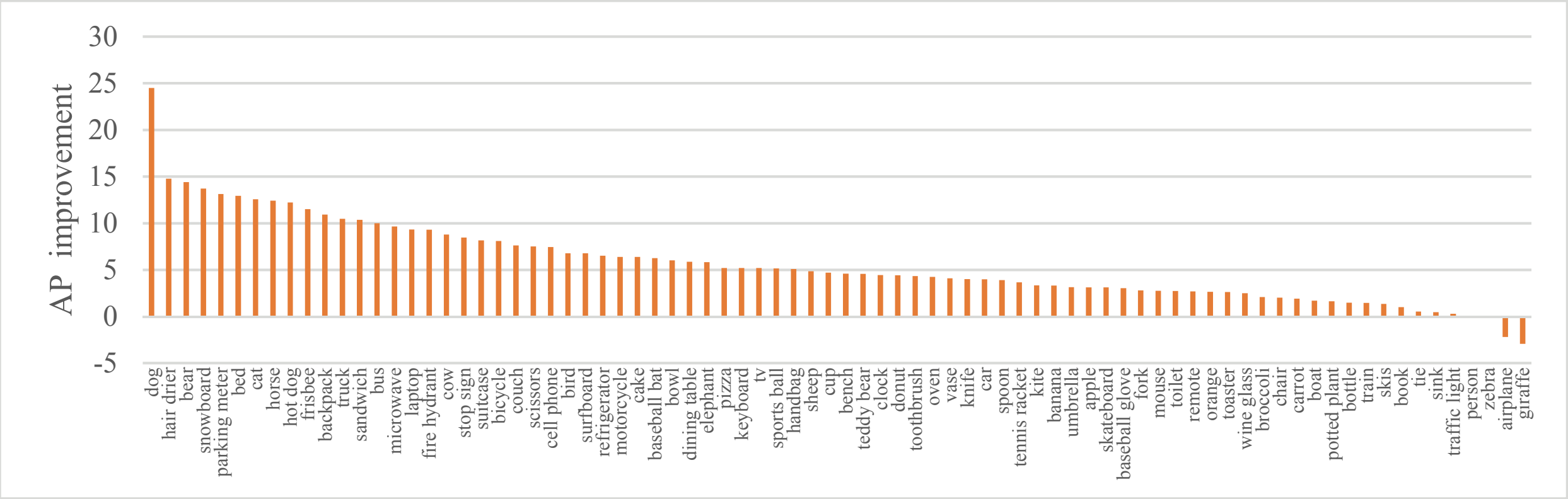

**Fig.6.** Average precision (AP)@0.5 improvement of each category for model 1) with OD-GCN framework.

Table.2. From these results, almost all AP evaluation metrics of categories have been improved by OD-GCN.

For the speed of OD-GCN, the 2-layer GCN and other new operations can be ignored compared with hundreds of former layers in the base model. So the total framework costs almost the same time as the base object detection model. During our experiments, the difference of the base model and OD-GCN on running time is less than 2%.

Finally, there are also some delightful visualized results shown in Fig.7, which proves that OD-GCN can reasonably adjust the wrong predictions to right ones. From these visualized results, it is obvious that OD-GCN can optimize the result of base detection model with understandable reasons, by surrounding objects' information with GCN.

## 5. CONCLUSION

Lack of utilization of surrounding information is a crucial issue for classical object detection frameworks. In this paper, we propose a novel framework named OD-GCN to solve the problem. OD-GCN provides a new way for object detection with GCN.

For the network structure, OD-GCN introduces a relationship knowledge graph and processes the graph with GCN. This knowledge graph is built by conditional probability between every two categories. We innovatively compress the raw confidence matrix as input feature for GCN. Also, we firstly try an adaptive parameter for classical graph convolution function.

For the experiment results, OD-GCN can improve detection performance of the several SSD and Faster R-CNN models at benchmark mAP evaluation metrics. Especially, the visualized results ensure that improvements are also quite reasonable in human's opinion, not only at benchmark evaluation metrics. As a flexible post-process framework, OD-GCN can help other object detection frameworks in the future.

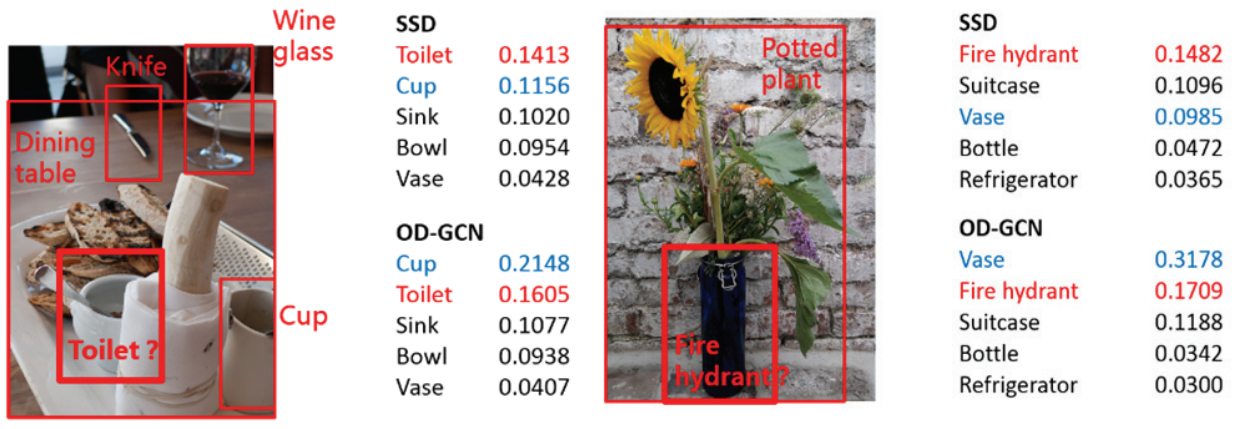

**Fig.7.** Two visualized reasonable adjustments of OD-GCN. OD-GCN successfully adjust toilet to cup with help of wine glass, dining table, cup and knife. It also successfully adjusts fire hydrant to vase with help of potted plant.